\RequirePackage[T1]{fontenc}
\documentclass[letterpaper, 10 pt, conference]{ieeeconf}
\IEEEoverridecommandlockouts
\usepackage{graphicx}
\usepackage{amsmath,amssymb}
\usepackage{booktabs,multirow,array}
\usepackage{cite}
\usepackage{algpseudocode}
\newcommand{\Kone}{K1}
\newcommand{\Ktwo}{K2}
\newcommand{\pihalf}{\pi_{0.5}}
\newcommand{\pizero}{\pi_0}
\newcommand{\meanstd}[2]{\ensuremath{#1{\scriptstyle\,\pm #2}}}
\title{\LARGE\bf GTI: Grounding Task Intent Throughout Long-Horizon VLA Manipulation}
\author{Tianqi Gong$^{*}$, Haochen Niu$^{*}$, Yusheng Zhao, Lantao Zhang,\\
Haoyu Liu, Rendong Ying, and Peilin Liu$^{\dagger}$\\[2pt]
{\normalsize Shanghai Jiao Tong University, China}\\[1pt]
{\footnotesize\texttt{\{gongtianqi,haochen\_niu,swagger,rdying,liupeilin\}@sjtu.edu.cn}}\\[2pt]
{\footnotesize $^{*}$Equal contribution\qquad $^{\dagger}$Corresponding author}
}
\begin{document}
\maketitle
\thispagestyle{empty}
\pagestyle{empty}

\begin{abstract}
Vision--language--action (VLA) models generate local actions from visual observations, language, and robot state, but long-horizon manipulation requires the global task intent to remain grounded in the currently executable context. We propose GTI (Grounding Task Intent), which uses the current observation and execution history to localize the active subtask and its execution phase. For the same local subtask, GTI generates two phase-dependent visual targets: an interaction/contact state and a local completion state. The local language instruction and the corresponding phase-specific visual target jointly condition the VLA, specifying both what to do now and what state should be reached at the current phase. During execution, anomaly monitoring and local recovery re-establish an intent-consistent execution context through rollback, re-observation, and phase re-evaluation after persistent deviation. We evaluate GTI on RoboCasa365, VLA-Arena Horizon-Long, and real-world PiPER tasks. Relative to matched Global-FT baselines, GR00T success on seen composite RoboCasa365 tasks increases from 14.6\%/12.1\% to 31.8\%/20.4\% under ID/OOD evaluation. On VLA-Arena, GTI+$\pihalf$ reaches 47.6\% overall success versus 45.2\% for an adapted ForeAct baseline, with gains of 5.2/6.0 percentage points on the harder L1/L2 tasks. In the three real-robot task experiments, mean full-task success increases from 21.7\% to 59.2\%. The results indicate that persistent semantic intent localization and phase-conditioned visual grounding improve multi-stage execution, while failure-triggered local recovery provides additional deployment-time support.
\end{abstract}

\section{INTRODUCTION}
Vision--language--action (VLA) models such as $\pihalf$~\cite{black05} and GR00T N1.5~\cite{groot} map visual observations, language, and robot state to actions, enabling multi-task manipulation. However, a global instruction remains unchanged while the robot encounters different subtasks and interaction states. Successful long-horizon execution therefore requires identifying not only which subtask is currently active, but also which interaction phase should be executed. Without this distinction, the policy may skip or repeat subtasks, switch phases prematurely, or continue executing after a failure.

Hierarchical methods narrow the gap between complex instructions and low-level actions. Hi Robot~\cite{hirobot} translates high-level reasoning into local language instructions, while ForeAct~\cite{foreact}, VISTA~\cite{vista}, and H-WM~\cite{hwm} use predicted future states as execution references. Language specifies what to do but leaves contact geometry to the executor, and a single completion state cannot represent the distinct interaction and completion relations within one subtask. For grasping or placing, establishing contact and reaching local completion require different visual targets within the same subtask. Phase therefore determines how intent should be concretized visually.

We therefore view long-horizon manipulation as a continuous task-intent-grounding problem with three coupled questions: which part of the global intent should be executed now; what visual target should represent that local intent in the current phase; and how an executable context consistent with that intent should be restored after persistent anomalies. GTI addresses these questions through history-based intent localization, phase-conditioned visual grounding, and failure-triggered local recovery.

GTI's planner localizes the active subtask and phase from current observations and execution history. A phase-conditioned visual predictor grounds the local instruction into an interaction/contact (\Kone{}) or completion (\Ktwo{}) target using the current observation, instruction, and phase. The target and instruction jointly condition the VLA. When an internal-feature monitor detects persistent anomalies, local rollback and re-observation are followed by phase re-evaluation to restore alignment with the current subtask and visual milestone.

The key contributions of this paper are summarized as follows:
\begin{itemize}
\item We introduce GTI, a continuous task-intent-grounding framework for long-horizon VLA manipulation that localizes the active subtask and phase from current observations and execution history, and re-establishes an intent-consistent context after persistent anomalies.
\item We propose a high-level mechanism combining a structured planner with phase-conditioned visual grounding: the planner localizes local task intent, while the visual model generates a \Kone{} interaction/contact target or \Ktwo{} completion target according to phase; both language and phase-specific visual targets condition the VLA.
\item We evaluate GTI on RoboCasa365, VLA-Arena Horizon-Long, and real PiPER tasks using complete-system comparisons, matched visual-grounding ablations, and deployment-sensitivity tests.
\end{itemize}

\begin{figure*}[t]
\centering\includegraphics[width=.714\textwidth]{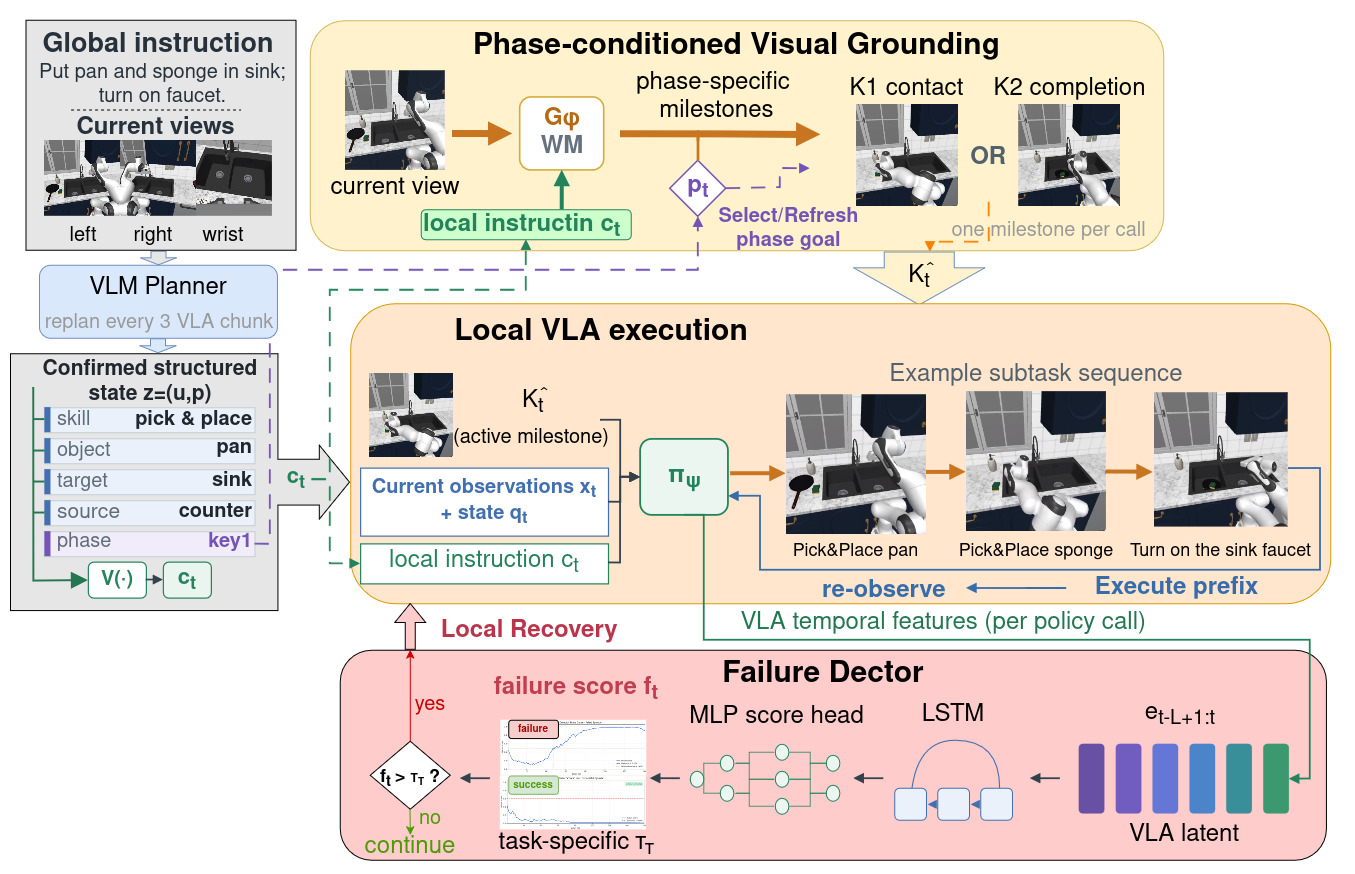}
\caption{Task-intent grounding throughout long-horizon VLA execution. The planner localizes the currently active intent, the visual predictor supplies phase-specific \Kone{}/\Ktwo{} milestones, and the VLA executes under the resulting language and visual conditions. Persistent anomalies trigger local recovery, followed by re-observation and intent re-grounding.}
\label{fig:overview}
\end{figure*}

\section{RELATED WORK}
\subsection{Generalist Robot Policies}
RT-2~\cite{rt2}, Open X-Embodiment~\cite{openx}, OpenVLA~\cite{openvla}, OpenVLA-OFT~\cite{oft}, $\pizero$~\cite{pi0}, $\pihalf$~\cite{black05}, and GR00T N1.5~\cite{groot} advance transfer, adaptation, and general-purpose action generation; feasible-action-neighborhood regularization improves VLA fine-tuning~\cite{fan}. RoboCasa365 provides a large-scale household benchmark~\cite{robocasa}. GTI adds long-horizon intent localization and phase-specific guidance.

\subsection{Hierarchical Task-Intent Localization}
Hi Robot~\cite{hirobot}, AtomicVLA~\cite{atomic}, hierarchical systems~\cite{multitask}, ReKep~\cite{rekep}, LoHo-Manip~\cite{loho}, LiLo-VLA~\cite{lilo}, and Long-VLA~\cite{longvla} provide high-level structure, relational constraints, progress-aware representations, or phase-aware perception. Unlike methods that modulate perception across motion phases, GTI uses observations and execution history to localize subtask and phase, then selects the corresponding task-semantic visual target.

\subsection{Visual Grounding Through Predicted Outcomes}
ForeAct~\cite{foreact}, VISTA~\cite{vista}, H-WM~\cite{hwm}, Reflective Planning~\cite{reflective}, Beyond Dense Futures/StructVLA~\cite{structvla}, V-JEPA 2~\cite{vjepa}, DreamVLA~\cite{dreamvla}, SKIP~\cite{skip}, and Fast-WAM~\cite{fastwam} use predicted states, world models, or sparse keyframes for guidance. Unlike event-based frames selected from gripper or kinematic changes, GTI supervises interaction/contact and completion targets from task semantics; the \Kone{}/\Ktwo{} milestones jointly condition action prediction with local language.

\subsection{Execution Anomalies and Intent Re-grounding}
Inner Monologue~\cite{inner}, AHA~\cite{aha}, and TCoT~\cite{tcot} use feedback for planning, failure detection, and recovery. GTI performs local rollback, re-observation, and phase re-evaluation while retaining the subtask intent, realigning execution with its \Kone{}/\Ktwo{} milestone.

\section{GTI FRAMEWORK}
As shown in Fig.~\ref{fig:overview}, GTI continuously grounds task intent into the currently executable context. A planner and execution history first localize the active subtask and phase to form a local language intent. A phase-conditioned visual predictor then grounds that intent into a visual milestone appropriate for the current phase. An interface-adapted VLA predicts actions from the local instruction and milestone. After persistent anomalies, local recovery rolls back and stabilizes the robot, then triggers re-observation, phase re-evaluation, and visual-target realignment. We represent the current guidance and execution mode as
\begin{equation} C_t=(c_t,\widehat K_t,\rho_t), \end{equation}
where $c_t$ is the local subtask instruction, $\widehat K_t$ is the current phase-specific visual milestone, and $\rho_t\in\{\mathrm{normal},\mathrm{recovery}\}$ is the execution mode. Given the current observation and robot state, the VLA is conditioned on $c_t$ and $\widehat K_t$; $\rho_t$ only selects native execution or local rollback/retry and is not an additional VLA input. The high-level planner is re-run at a fixed interval to update the confirmed subtask and phase from the latest observation.

\subsection{Memory-Based Intent Localization}
The planner receives the global task intent $g$, current views $x_t=\{I_t^L,I_t^R,I_t^W\}$, and history $m_t$. The global instruction and the ordered history of completed subtasks form the text context, augmented with five field queries: skill, object, target, source, and phase. The execution history provides temporal context for distinguishing visually similar states at different stages of a task, allowing the planner to identify the next subtask without relying solely on the current observation. Training histories use ground-truth execution sequences. At runtime, a semantic switch requires two consecutive confirmations; a new subtask must remain stable for one additional call before it becomes history-eligible and is appended when the next semantic switch is confirmed. Phase updates are not written to history.

Let $f\in\{s,o,d,r,p\}$ denote skill, object, target location, source, and phase, respectively. A learnable query $q_f$ produces
\begin{equation}h_f=H_\theta(x_t,g,m_t,q_f).\end{equation}
We use field-parallel token decoding over the shared Qwen vocabulary $\mathcal V$. Each field contains at most $T_{\max}=16$ positions including EOS. At position $j$, a learnable field-and-position embedding $e_{f,j}$ is inserted before the shared LM head:
\begin{equation}
\begin{aligned}
\ell_{f,j}&=W_{\rm LM}(h_f+e_{f,j}),\\
\mathcal L_P&=\frac{\sum_f\sum_{j=1}^{T_f}
\mathrm{CE}(\ell_{f,j},y_{f,j})}{\sum_fT_f}.
\end{aligned}
\end{equation}
Padding positions are excluded. All fields share the LM head and vocabulary, while token positions are independently supervised and decoded in parallel. Inference uses neither autoregressive beam search nor a finite label list:
\begin{equation}
\begin{aligned}
& p_{f,j}(v) = \mathrm{Softmax}(\ell_{f,j})_v,\\
& \widehat y_{f,j} = \operatorname*{arg\,max}_{v\in\mathcal V}\,p_{f,j}(v),\\
& J_f = \min\{j:\widehat y_{f,j}=\mathrm{EOS}\}-1,\\
& \widehat y_f = \widehat y_{f,1:J_f}.
\end{aligned}
\end{equation}
If EOS is absent, $J_f=T_{\max}$, yielding an implicit non-enumerated string space.

We write the structured state as $z=(u,p)$, where $u=(s,o,d,r)$ contains semantic fields and $p$ is phase. When semantics are unchanged, an allowable phase transition can be accepted from one planner call; normal execution permits only a transition from \Kone{} to \Ktwo{}. Upon advancement to \Ktwo{}, the \Ktwo{} milestone is regenerated while \Kone{} is retained; a semantic switch clears the visual-target cache. Post-recovery phase re-evaluation and cache reuse are described in Sec.~III-C. A deterministic template $V(\cdot)$ converts task-relevant semantic fields into the local instruction $c_t$ according to the current skill, while $p$ is separately passed to the phase-conditioned visual-grounding module.

\subsection{Phase-Conditioned Visual Grounding and Executor Adaptation}
Intent localization determines what to execute, but the same local intent can require different outcome references at different interaction phases. We therefore adapt a pretrained video generator as a phase-conditioned visual-grounding model. Given current observation $I_t^L$, local instruction $c_t$, and confirmed phase $p_t$, it predicts
\begin{equation*}\widehat K_t=G_\phi(I_t^L,c_t,p_t).\end{equation*}
Phase is an explicit conditioning variable during training and inference. At the manipulation level, \Kone{} is the first state in which the current atomic operation establishes the task-relevant robot--object or object--environment interaction relation; \Ktwo{} is the first state in which the local subtask completion predicate becomes true. In RoboCasa and VLA-Arena, simulator states, action signals, and task predicates provide \Kone{}/\Ktwo{} milestone labels and phase labels, associating the interaction and completion phases of the same subtask with different visual outcomes. The confirmed phase is concatenated directly with the local instruction as Wan text conditioning, e.g., ``pick up the apple and place it in the bowl; current phase: \Kone.'' For \Kone{} pairs, the current frame is sampled before the contact milestone; for \Ktwo{} pairs, it is sampled after \Kone{} and before \Ktwo. We use at most three current frames per target and require at least a 20-frame current-to-target gap.

\begin{figure*}[t]
\centering\includegraphics[width=.70\textwidth]{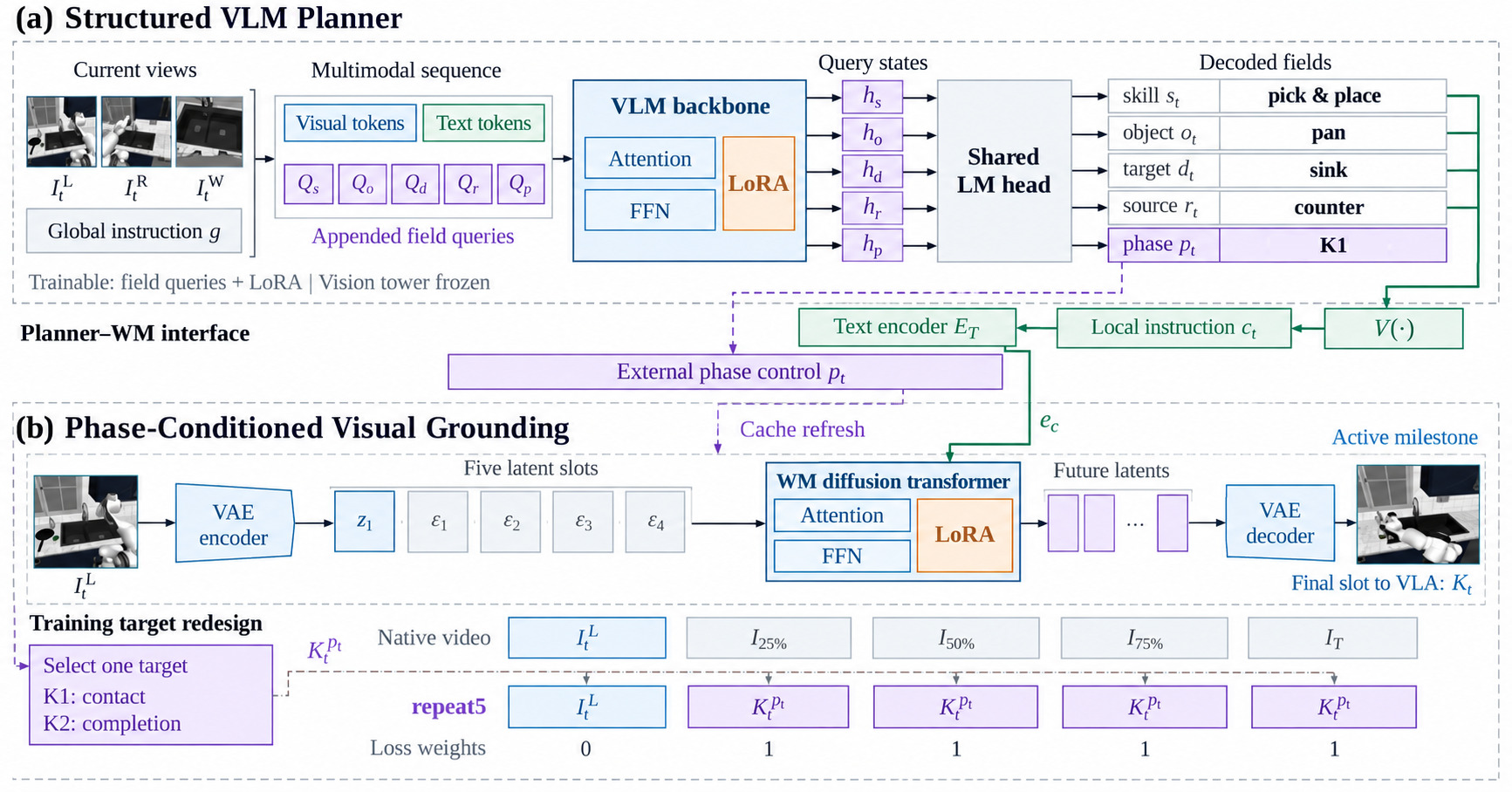}
\caption{Adaptation of intent localization and phase-conditioned visual grounding. Phase controls \Kone{}/\Ktwo{} generation; five RGB supervision frames map to two VAE temporal latent positions.}
\label{fig:adaptation}
\end{figure*}

During training, to preserve Wan's native temporal video interface without changing its architecture, we use repeated-goal supervision. With VAE temporal compression ratio 4, each training sample contains one current observation followed by four copies of the same phase-specific image goal. For milestone $K=K_t^{p_t}$, the five-frame sequence and its RGB-frame weights are
\begin{equation}
\begin{aligned}
V_t &= [I_t^L,K,K,K,K],\\
w &= [0,1,1,1,1],
\end{aligned}
\end{equation}
where $V_t\in\mathbb R^{3\times5\times256\times256}$.
Let $E$ be the VAE encoder and $\sigma$ the mixing coefficient. Flow-matching velocity prediction uses
  \begin{equation}
  \begin{aligned}
  z_0&=E(V_t),\qquad \epsilon\sim\mathcal N(0,I),\\
  z_\sigma&=(1-\sigma)z_0+\sigma\epsilon,\qquad v^*=\epsilon-z_0.
  \end{aligned}
  \end{equation}
  with
  \begin{equation}
  \mathcal L_{\rm FM}=
  \frac{\sum_iM_i\|v_\phi(z_\sigma,\sigma,c_t,p_t)_i-v_i^*\|_2^2}
  {C\sum_iM_i}.
  \end{equation}

The first frame is conditioning only, while the four future positions are supervised by the same goal $K$. This repetition is used only for training to preserve the pretrained temporal interface. At inference, the target latent positions are initialized from noise and generated under the current observation, local instruction, and confirmed phase; the final decoded frame is used as $\widehat K_t$. Here $i$ indexes temporal/spatial latent positions and $C$ is the number of latent channels. RGB-frame weights are interpolated to latent temporal length and combined with the first-frame conditioning mask to obtain $M$, so loss is applied only to target latent positions. The milestone is refreshed at initialization, confirmed semantic switches, and transitions from \Kone{} to \Ktwo{}; otherwise the current instruction and cached milestone are reused.

The low-level VLA is adapted to condition directly on the local instruction and predicted image while retaining its native action parameterization. For executor $\pi_\psi$,
\begin{equation}a_{t:t+H-1}=\pi_\psi(x_t,q_t,c_t,\widehat K_t),\end{equation}
where $q_t$ is robot state and $H$ the action horizon. Between high-level updates, the executor reuses the current instruction and cached milestone. For all three executors, $\widehat K_t$ is an additional fourth visual input processed together with physical camera views by the executor's shared visual encoder: Eagle for GR00T and PaliGemma for $\pizero$/$\pihalf$. No goal-image-specific encoder or separate visual branch is introduced. The adaptation pipeline is summarized in Fig.~\ref{fig:adaptation}.

\subsection{Anomaly Monitoring and Failure-Triggered Re-grounding}
\textbf{Anomaly monitoring.} GTI uses an LSTM--MLP (1024, 256, single-layer LSTM-256, MLP, dropout 0.1) to model temporal patterns in final denoised VLA action latents and predict anomaly probabilities. Let $e_{i,1:t}$ denote the latent sequence of rollout $i$ up to chunk $t$, with $p_{i,t}=\sigma(f_\omega(e_{i,1:t}))$. For a failed rollout, $t_i^*$ denotes the onset of the terminal persistent-anomaly suffix. Define
\begin{equation*}
\widetilde A_{i,t}=\sum_{k=t_i^*}^{t}m_{i,k}p_{i,k},\qquad
\widetilde n_{i,t}=\sum_{k=t_i^*}^{t}m_{i,k}.
\end{equation*}
The training objective is
  \begin{equation}
  \begin{aligned}
  \mathcal L_F=\frac1Z\bigg[&
  \sum_{i,t<t_i^*}m_{i,t}p_{i,t}+\sum_{i,t\ge t_i^*}
  m_{i,t}(\widetilde n_{i,t}-\widetilde A_{i,t})\bigg].
  \end{aligned}
  \end{equation}
where $m_{i,t}\in\{0,1\}$ masks uncertain chunks from both direct and cumulative terms and $Z=\sum_{i,t}m_{i,t}$. Since $p_{i,t}\in[0,1]$, the cumulative gap is nonnegative and needs no hinge. The cumulative suffix deliberately gives early chunks near anomaly onset greater weight because they contribute to more subsequent prefixes. Longer persistent failures therefore receive larger total weight; we retain this dependence instead of normalizing by suffix length. For a successful rollout ending at chunk $T_i$, we set $t_i^*=T_i+1$, leaving only the first term.

\textbf{Failure-triggered local re-grounding.} A persistent anomaly triggers local rollback, stabilization, and cooldown. The current subtask and \Kone{}/\Ktwo{} cache are retained, and phase is re-evaluated from the new observation. A cached milestone is reused when available; otherwise it is regenerated. A transition from \Ktwo{} back to \Kone{} is allowed only after recovery; the subsequent advancement to \Ktwo{} regenerates its milestone. Online recovery is triggered when the anomaly score exceeds the calibrated dynamic threshold for two consecutive policy chunks. Failure detection uses internal VLA action-generation features rather than visual-prediction error.

\begin{figure}[tb]
\footnotesize
\hrule\vspace{2pt}
\noindent\textbf{Algorithm 1: GTI Task-Intent Grounding and Execution}\par
\vspace{1pt}
\begin{algorithmic}[1]
\Require $g$, $\Delta$, $N_c$, $\lambda_p,\lambda_v,\lambda_r$
\State Initialize $z,c,K\gets\varnothing$, $\mathcal K\gets\{\}$,
$m\gets[]$, $\rho\gets\mathrm{normal}$, $b,e\gets\mathrm{false}$;
initialize recovery eligibility.
\While{episode not terminated}
    \State Observe $(x_t,q_t)$; \textbf{if} $\lambda_p=0$, set
    $c\gets g$, $K\gets0_{\rm img}$;
    \textbf{if} $b$, enable post-recovery phase reselection.
    \If{$\lambda_p=1$ and ($z=\varnothing$ or $b$ or $\Delta$ reached)}
        \State $(z^{\rm old},e^{\rm old})\gets(z,e)$;
        $\widetilde z=(\widetilde u,\widetilde p)\gets P_\theta(x_t,g,m)$;
        $(u,\chi)\gets\mathrm{ConfirmStable}_{N_c}(\widetilde u)$.
        \State $\eta\gets0$;
        \textbf{if} $\chi$, $p\gets\widetilde p$;
        \textbf{else if} $u$ unchanged and $\widetilde p$ allowable,
        $\eta\gets[p\ne\widetilde p]$, $p\gets\widetilde p$;
        $z\gets(u,p)$.
        \State \textbf{if} $\chi$, append eligible $z^{\rm old}$ to $m$,
        clear $\mathcal K$, reset subtask recovery eligibility;
        \textbf{else if} semantics unchanged, $e\gets\mathrm{true}$.
        \State \textbf{if} $\chi$ or $\eta$ or $b$, $c\gets V(z)$;
        \textbf{if} $\lambda_v=1$, reuse/generate $K\gets\mathcal K[p]$;
        \textbf{else}, $K\gets0_{\rm img}$; $b\gets\mathrm{false}$.
    \EndIf
    \State \textbf{if} $c=\varnothing$ or
    ($\lambda_p=1$ and $z=\varnothing$) or
    ($\lambda_v=1$ and $K=\varnothing$), \textbf{continue}.
    \State Execute native VLA chunk from $(x_t,q_t,c,K)$;
    update monitor/cooldown.
    \State \textbf{if} $\lambda_r=1$ and recovery allowed and alarm
    persists, $\rho\gets\mathrm{recovery}$;
    rollback/stabilize retaining $(z,c,\mathcal K,m)$;
    mark recovery used; $b\gets\mathrm{true}$;
    $\rho\gets\mathrm{normal}$.
\EndWhile
\end{algorithmic}
\vspace{2pt}\hrule
\end{figure}

\section{EXPERIMENTS}
\subsection{Experimental Setup}
\textbf{Base models.} We use Qwen2.5-VL-3B-Instruct~\cite{qwen} for structured intent localization and Wan2.2-TI2V-5B-Diffusers~\cite{wan} for phase-conditioned visual grounding. We adapt different VLA executors to assess the generality of GTI while retaining their native action parameterizations.

\textbf{Benchmarks and splits.} RoboCasa365~\cite{robocasa} contains 18 seen atomic, 16 seen composite, and 16 unseen composite tasks with ID and cross-scene OOD evaluation; adaptation uses seen-task trajectories only. VLA-Arena Horizon-Long~\cite{vlaarena} uses L0+L1 for adaptation and holds out L2. We denote L0 as Atomic and L1/L2 as Composite to align hierarchical statistics with RoboCasa; L1 contains in-distribution multi-stage tasks and L2 contains harder held-out tasks. Real-robot evaluation uses an AgileX PiPER 6R on cabinet search/place, bell-pepper--banana swapping, and kiwi search/place.

\textbf{Comparisons.} We compare GTI with matched Global-FT baselines on all three platforms and with adapted ForeAct~\cite{foreact} on VLA-Arena. Component studies are discussed with the corresponding benchmark results.

\textbf{Evaluation metrics.} Full-system RoboCasa and VLA-Arena results are averaged over seeds 7, 42, and 142; the RoboCasa deployment-component test uses seed 7. RoboCasa evaluates each task over 50 trials per seed. Its overall success is the task-count-weighted average of seen atomic (18 tasks), seen composite (16), and unseen composite (16) success rates. VLA-Arena evaluates 50 episodes per task and seed (500, 250, and 250 episodes for L0, L1, and L2, respectively) with matched \texttt{episode\_idx}; its overall success weights L0, L1, and L2 by 0.50, 0.25, and 0.25. Each real-robot condition contains 30 trials.

\subsection{Implementation Details}
\textbf{Planner adaptation.} Simulation and real-robot planners use the same Qwen2.5-VL-3B setup: frozen visual encoder, LoRA rank 16, learning rate $10^{-4}$, and 2 epochs, with shared atomic/composite sampling and denser sampling near the transition from \Kone{} to \Ktwo{}. Data sources differ: RoboCasa uses 89,260 atomic and 215,934 composite labeled states; VLA-Arena derives training data from 728 L0/L1 demonstrations; the real robot uses 1,000 successful teleoperated demonstrations across the three tasks.

\textbf{Visual grounding and goal conditioning.} All settings use the same pretrained Wan and LoRA hyperparameters (rank 16, learning rate $10^{-4}$), with dataset-dependent training length. RoboCasa and VLA-Arena use 148,790 and 4,317 \Kone{}/\Ktwo{} records for 4 and 12 epochs; the real robot constructs 3,000 \Kone{}/\Ktwo{} records from the same 1,000 successful demonstrations and trains for 12 epochs. Visual-goal variants use local instruction+goal image; Planner+phase uses local instruction+explicit interaction/completion text and no goal image.

\textbf{RoboCasa executors.} GR00T N1.5 and $\pihalf$ Global-FT/GTI-FT start from identical checkpoints and use the same seen trajectories, views, and fine-tuning budget. Global-FT uses the global instruction; GTI-FT uses local instruction plus a phase-specific milestone. Interface adaptation uses one GPU, per-device batch 8, gradient accumulation 4 (effective batch 32), and 20,000 updates.

\textbf{VLA-Arena executors.} Planner and visual models train on L0+L1. Five matched $\pihalf$ variants train one epoch on 728 trajectories (478 L0, 250 L1; 115,198 frames), sharing LoRA initialization, normalization, action supervision, optimizer, and batch size 8. Planner language uses the local subtask; Planner+phase additionally provides interaction/completion text but no goal image; Single-\Ktwo{} and \Kone{}/\Ktwo{} use a completion goal and phase-specific visual goals. Phase text is predicted online by the same planner/confirmation logic as GTI, not from ground-truth phase labels. Recovery is disabled for all matched variants and enabled only in GTI.

\textbf{Real robot and ForeAct.} The three PiPER tasks use 400/300/300 successful demonstrations. Global-FT and GTI-FT use the same demonstrations for 15,000 updates, and subtask boundaries and \Kone{}/\Ktwo{} phase keyframes are manually annotated. ForeAct is adapted on the same L0+L1 data; its Qwen3-VL planner remains frozen, the visual predictor is initialized from the public VisualForesight checkpoint, and only the Sana Transformer in ImGen is updated.

\textbf{Anomaly-monitor training and calibration.} Simulation and real-robot monitors use the same labels and train for 30 epochs: successful chunks are on-track; 48 consecutive low-level steps without predicate/state/action progress are suspicious; terminal predicate failure is failed; chunks near anomaly onset are uncertain. RoboCasa uses 1,000 independent rollouts (248 success/752 failed), and VLA-Arena uses 1,000 rollouts (399/601). For the real robot, the 1,000 successful rollouts are split at the episode level into two disjoint subsets of 500. One subset is combined with 500 additionally collected failure rollouts to train the anomaly monitor, while the other successful subset is excluded from monitor parameter training and used only for functional conformal calibration after the monitor is frozen. Calibration uses only successful rollouts and targets 5\% episode-level FPR with rank $\lceil(N+1)(1-\alpha)\rceil$, clipped to $[1,N]$, $\alpha=0.05$, and threshold $\tau_t=\mu_t+\widehat q\sigma_t$, where $\mu_t$ and $\sigma_t$ are the standardization statistics at chunk $t$. All thresholds are fixed before evaluation and are not recalibrated during testing.

\textbf{Execution and recovery.} RoboCasa replans every three chunks. GR00T uses four flow-matching steps to predict/execute $H=h=16$ actions; at 20 Hz each chunk lasts 0.8 s and the planning-check interval is 2.4 s, excluding inference pauses. VLA-Arena uses 8-action chunks, replans every 24 low-level steps, and caps episodes at 300 steps. Semantic switches require two calls for confirmation; startup queries are exempt. All three platforms use an 8-step $+z$ rollback, 4-step stabilization, and 60-step cooldown, with at most one recovery per subtask. Simulator state is never reset. Post-recovery phase re-evaluation and milestone reuse follow Sec.~III-C.

\begin{table}[t]
\caption{{\normalfont Offline intent-localization accuracy (\%). E4/E5: Exact-4/5; $s/o/d/r/p$: skill/object/target/source/phase. VLA-Arena Atomic=L0 and Composite=L1/L2.}}
\label{tab:localization}\centering
\setlength{\tabcolsep}{2.35pt}
\renewcommand{\arraystretch}{1.12}
\begin{tabular}{@{}l*{7}{>{\footnotesize}c}@{}}\toprule
Split&E4&E5&$s$&$o$&$d$&$r$&$p$\\\midrule
RoboCasa Overall&90.30&79.55&95.98&91.83&93.48&95.81&83.87\\
\quad Atomic&96.44&90.30&100.00&97.78&98.67&100.00&93.68\\
\quad Composite&84.73&69.79&92.23&86.43&88.77&92.00&74.96\\
\midrule
VLA-Arena Overall&85.13&83.36&85.54&86.08&85.68&86.36&94.95\\
\quad Atomic (L0)&99.18&96.73&99.59&100.00&99.59&100.00&97.55\\
\quad Composite (L1/L2)&78.07&76.64&78.48&79.10&78.69&79.51&93.65\\\bottomrule
\end{tabular}
\end{table}

\begin{table}[t]
\caption{{\normalfont Offline phase-conditioned visual grounding with ground-truth local instructions. PSNR/SNR: dB.}}
\label{tab:imagequality}\centering
\begin{tabular}{llccc}\toprule
Benchmark&Split&MSE$\downarrow$&PSNR$\uparrow$&SNR$\uparrow$\\\midrule
RoboCasa&Atomic&0.0847&18.12&7.43\\
&Composite&0.0921&17.57&6.63\\
\midrule
VLA-Arena HL&L0&0.0479&19.53&3.44\\
&L1&0.0431&20.25&3.87\\\bottomrule
\end{tabular}
\end{table}

\subsection{Benchmark Results}
\textbf{(1) System decision accuracy.} Table~\ref{tab:localization} evaluates offline localization of the active intent, using histories rolled out from the planner's preceding predictions. Exact-4 matches $(s,o,d,r)$ and Exact-5 additionally requires phase. VLA-Arena L0 is reported as Atomic and L1/L2 as Composite. Atomic localization is markedly easier because composite tasks require identifying both the current subtask and phase from observations and execution history.

\begin{figure*}[t]
\centering\includegraphics[width=.82\textwidth]{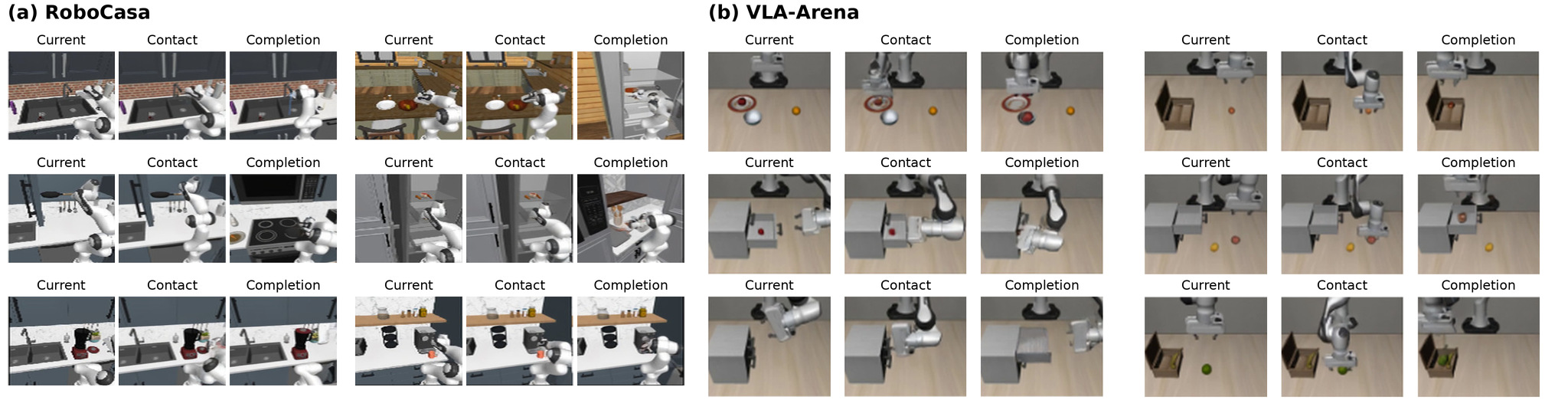}
\caption{Phase-conditioned visual grounding on RoboCasa and VLA-Arena: current observation, \Kone{} interaction/contact milestone, and \Ktwo{} completion milestone.}
\label{fig:visualgoals}
\end{figure*}

Table~\ref{tab:imagequality} reports low-level image quality, which does not directly measure phase semantics or interaction relations. We therefore manually diagnose task-stratified random samples from held-out evaluation trajectories: 10 prediction groups for each of 50 RoboCasa and 20 VLA-Arena tasks (700 total). \Kone{} is correct when the gripper forms reasonable contact with the correct target; \Ktwo{} when the predicted state satisfies the local completion condition; object identity and spatial placement are also checked. Correct \Kone{}/\Ktwo{}/geometry rates are 631/700 (90.1\%), 581/700 (83.0\%), and 504/700 (72.0\%). Failures are mainly phase confusion or local blur and concentrate on RoboCasa Composite-Unseen and VLA-Arena L2, consistent with their more limited control gains. Representative predictions are shown in Fig.~\ref{fig:visualgoals}.

\begin{table}[t]
\caption{{\normalfont VLA-Arena success (\%, mean$\pm$std.; seeds 7/42/142). L0+L1 are adapted; L2 is held out. $\pi_{0.5}$/$\pi_0$ denote Global-FT baselines; ForeAct uses K2-FT (local language+K2). GTI additionally uses phase grounding and local re-grounding.}}
\label{tab:vlaarena}\centering
\small
\setlength{\tabcolsep}{2.5pt}
\renewcommand{\arraystretch}{1.12}
\begin{tabular}{@{}lcccc@{}}\toprule
Model&L0&L1&L2&Overall\\\midrule
$\pihalf$&\meanstd{85.8}{2.4}&\meanstd{3.5}{1.5}&\meanstd{0.0}{0.0}&\meanstd{43.8}{0.9}\\
ForeAct+$\pihalf$&\meanstd{87.3}{2.1}&\meanstd{6.3}{0.8}&\meanstd{0.0}{0.0}&\meanstd{45.2}{1.2}\\
GTI+$\pihalf$&\meanstd{86.4}{1.1}&\meanstd{11.5}{0.6}&\meanstd{6.0}{0.4}&\meanstd{47.6}{0.7}\\
$\Delta$ (GTI$-\pi_{0.5}$)&+0.6&+8.0&+6.0&+3.8\\\midrule
$\pizero$&\meanstd{79.5}{1.4}&\meanstd{9.9}{1.2}&\meanstd{0.0}{0.0}&\meanstd{42.2}{0.9}\\
ForeAct+$\pizero$&\meanstd{80.2}{1.2}&\meanstd{12.2}{1.2}&\meanstd{1.7}{0.8}&\meanstd{43.6}{1.1}\\
GTI+$\pizero$&\meanstd{80.7}{1.8}&\meanstd{16.7}{0.6}&\meanstd{3.2}{1.1}&\meanstd{45.3}{1.0}\\
$\Delta$ (GTI$-\pi_0$)&+1.2&+6.8&+3.2&+3.1\\\bottomrule
\end{tabular}
\end{table}

\begin{table}[t]
\caption{{\normalfont $\pihalf$ VLA-Arena ablation (\%, mean$\pm$std.; seeds 7/42/142).
Matched L0+L1 data/budget; L2 held out. Phase text uses the online planner
phase and no goal image. Recovery is disabled for matched variants and
enabled only in GTI.}}
\label{tab:ablation}
\centering
\footnotesize
\setlength{\tabcolsep}{3.0pt}
\renewcommand{\arraystretch}{1.12}

\begin{tabular}{@{}>{\raggedright\arraybackslash}p{.38\columnwidth}cccc@{}}
\toprule
Configuration & L0 & L1 & L2 & Overall\\
\midrule
Global instruction
& \meanstd{85.8}{2.4} & \meanstd{3.5}{1.5} & \meanstd{0.0}{0.0} & \meanstd{43.8}{0.9}\\
Planner language
& \meanstd{86.5}{2.0} & \meanstd{4.1}{0.8} & \meanstd{0.0}{0.0} & \meanstd{44.3}{1.1}\\
Planner+phase text (no goal image)
& \meanstd{86.3}{1.2} & \meanstd{5.6}{0.3} & \meanstd{1.1}{0.5} & \meanstd{44.8}{0.7}\\
Planner+Single-\Ktwo
& \meanstd{86.0}{0.9} & \meanstd{6.0}{1.0} & \meanstd{1.7}{0.8} & \meanstd{44.9}{0.7}\\
\Kone{}/\Ktwo{} w/o recovery
& \meanstd{86.8}{0.7} & \meanstd{9.6}{0.8} & \meanstd{4.7}{0.4} & \meanstd{47.0}{0.5}\\
GTI
& \meanstd{86.4}{1.1} & \meanstd{11.5}{0.6} & \meanstd{6.0}{0.4} & \meanstd{47.6}{0.7}\\
\bottomrule
\end{tabular}
\end{table}

\textbf{(2) Performance on VLA-Arena.} Relative to the corresponding $\pihalf$ and $\pizero$ baselines in Table~\ref{tab:vlaarena}, GTI improves overall success by 3.8 and 3.1 percentage points, respectively, with most gains on L1/L2; gains over adapted ForeAct are likewise concentrated on multi-stage tasks.

Table~\ref{tab:ablation} isolates planner language, phase text without a goal image, a single \Ktwo{} goal, and phase-conditioned \Kone{}/\Ktwo{} goals under matched $\pihalf$ adaptation; recovery is enabled only for GTI. Planner language increases overall success from 43.8\% to 44.3\%; phase text reaches 44.8\%, and Single-\Ktwo{} 44.9\%. \Kone{}/\Ktwo{} without recovery reaches 47.0\%, exceeding phase text by 2.2/4.0/3.6 points and Single-\Ktwo{} by 2.1/3.6/3.0 points on Overall/L1/L2. The result indicates complementary spatial and interaction information beyond a phase token or single completion goal. It does not increase average Wan calls (L0: 2.00/1.99; L1: 11.64/10.60). Local re-grounding further raises success to 47.6\%, mainly on L1/L2, though L2 remains low.

\begin{table*}[!t]
\centering
\caption{{\normalfont RoboCasa systems (\%, mean $\pm$ std.; seeds 7/42/142; 50 trials/task/seed). Global-FT uses global instructions; GTI uses GTI-FT plus all online modules. $\Delta$: absolute gain over matched Global-FT (percentage points).}}
\label{tab:robocasa}
\setlength{\tabcolsep}{5.2pt}
\begin{tabular}{lcccccccc}
\toprule
& \multicolumn{4}{c}{ID} & \multicolumn{4}{c}{OOD}\\
\cmidrule(lr){2-5}\cmidrule(lr){6-9}
System / executor & A & CS & CU & Overall ($\Delta$) & A & CS & CU & Overall ($\Delta$)\\
\midrule
GR00T Global-FT & $45.9\pm1.0$ & $14.6\pm2.9$ & $3.4\pm1.1$ & $22.3\pm0.2$ & $45.9\pm1.9$ & $12.1\pm1.0$ & $2.3\pm1.5$ & $21.1\pm0.6$\\
GTI + GR00T & $52.8\pm0.9$ & $31.8\pm1.7$ & $5.2\pm1.3$ & $30.8\pm1.1$ (+8.5) & $44.5\pm2.1$ & $20.4\pm0.9$ & $4.2\pm0.9$ & $23.9\pm1.3$ (+2.8)\\
\midrule
$\pi_{0.5}$ Global-FT & $48.8\pm2.7$ & $8.3\pm0.6$ & $2.0\pm0.3$ & $20.9\pm1.0$ & $50.7\pm1.6$ & $7.5\pm0.4$ & $1.8\pm0.6$ & $21.2\pm0.6$\\
GTI + $\pi_{0.5}$ & $48.4\pm1.6$ & $17.6\pm2.7$ & $5.3\pm1.1$ & $24.8\pm1.7$ (+3.9) & $49.0\pm1.2$ & $12.0\pm0.9$ & $4.8\pm0.9$ & $23.0\pm1.0$ (+1.8)\\
\bottomrule
\end{tabular}
\end{table*}

\begin{table*}[!t]
\centering
\caption{{\normalfont RoboCasa deployment-component test, seed 7 (\%). A/CS/CU: seen atomic/composite and unseen composite; P/WM/R: planner/visual model/recovery. One GR00T GTI-FT checkpoint is shared. With WM removed, the fourth goal view is an all-zero image; with P removed, planner-dependent state checks are bypassed and the executor uses the global instruction with the zero goal view.}}
\label{tab:deployment}
\setlength{\tabcolsep}{8pt}
\begin{tabular}{lcccccccc}
\toprule
& \multicolumn{4}{c}{ID} & \multicolumn{4}{c}{OOD}\\
\cmidrule(lr){2-5}\cmidrule(lr){6-9}
Deployment configuration & A & CS & CU & Overall & A & CS & CU & Overall\\
\midrule
GTI w/o P, WM, R & 51.1 & 24.1 & 3.8 & 27.3 & 46.1 & 16.1 & 1.4 & 22.2\\
GTI w/o WM, R & 48.9 & 28.6 & 3.6 & 27.9 & 42.9 & 18.4 & 4.1 & 22.6\\
GTI w/o R & 53.7 & 32.0 & 5.4 & 31.3 & 47.3 & 20.1 & 4.0 & 24.7\\
GTI & 53.4 & 33.3 & 6.7 & 32.0 & 46.7 & 21.4 & 5.2 & 25.3\\
\bottomrule
\end{tabular}
\end{table*}

\textbf{(3) Performance on RoboCasa.} Table~\ref{tab:robocasa} compares separately adapted full systems. Relative to matched Global-FT, GR00T gains 8.5/2.8 points in ID/OOD overall success and $\pihalf$ gains 3.9/1.8. OOD gains are smaller, indicating that performance remains constrained by executor generalization.

Table~\ref{tab:deployment} compares four configurations, progressively enabling the planner, visual model, and recovery using the same seed-7 GR00T GTI-FT checkpoint without retraining. Thus, the test measures deployment sensitivity rather than strict causal ablation. The GTI configuration achieves the highest ID/OOD overall success; removing visual grounding or recovery lowers overall performance, while planner removal is more task dependent. On seed 7, the monitor reaches 98.11\% subtask precision, 67.7\% recall, and 9.8\% FPR; the latter is a subtask-level rate and is not directly comparable with the 5\% episode-level conformal target.

\subsection{Real-World Deployment}
\begin{figure*}[!t]
\centering
\includegraphics[width=.98\textwidth]{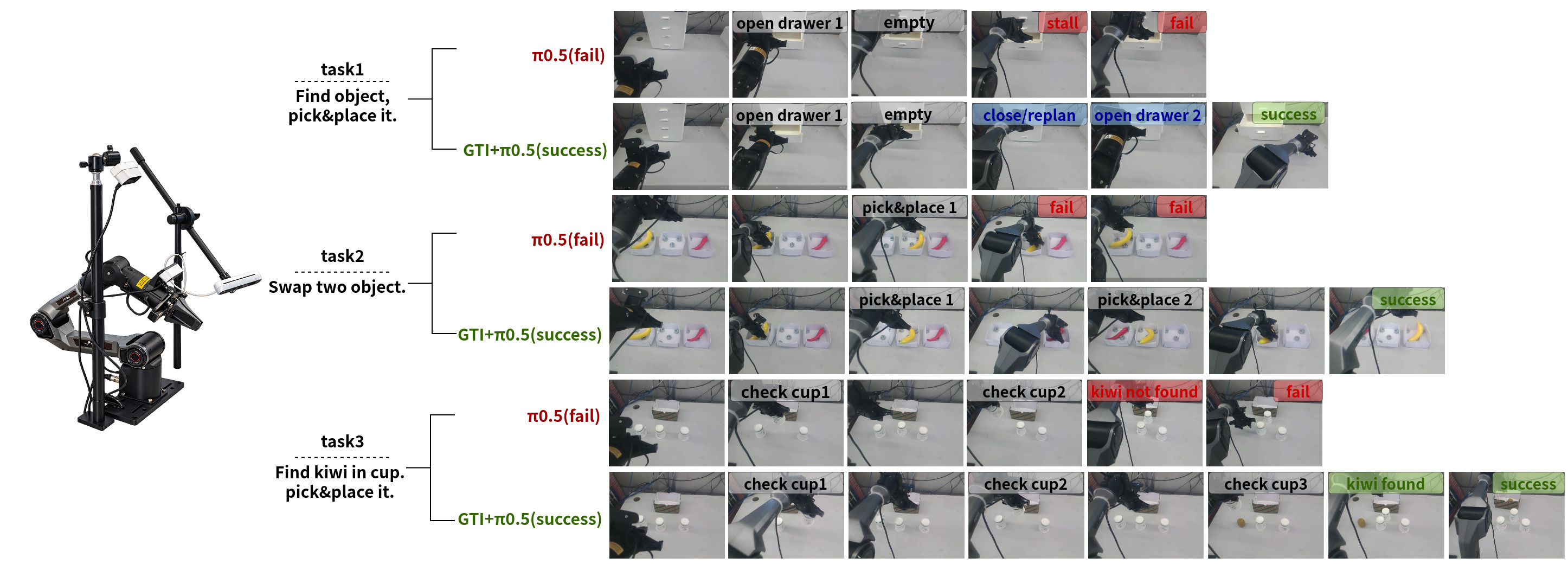}
\caption{Real-robot evaluation. Left: rollout keyframes for cabinet search, bell-pepper--banana swapping, and kiwi search/place (top: Global-FT; bottom: GTI). Right: AgileX PiPER 6R platform.}
\label{fig:rollouts}
\end{figure*}

\begin{table*}[!t]
\caption{{\normalfont Real-robot stage counts/full-task success (30 trials/condition). Task 1-A: S1/S2=open/grasp-and-place; 1-B: S1--S4=open 1/close 1/open 2/grasp-and-place. Full success ends at A-S2, B-S4, and S3 for Tasks 2--3; $\Delta$ is gain over Global-FT (percentage points); ``--'' is N/A.}}
\label{tab:real}\centering\setlength{\tabcolsep}{2.8pt}
\begin{tabular}{llccccc|cccc|cccc}
    \toprule
    & & \multicolumn{5}{c}{Task 1} & \multicolumn{4}{c}{Task 2} & \multicolumn{4}{c}{Task 3}\\
    \cmidrule(lr){3-7}\cmidrule(lr){8-11}\cmidrule(lr){12-15}
    Method & Cond. & S1 & S2 & S3 & S4 & Success ($\Delta$) & S1 & S2 & S3 & Success ($\Delta$) & S1 & S2 & S3 & Success ($\Delta$)\\\midrule
$\pihalf$ Global-FT&A&25&13&--&--&43.3\%&22&15&4&13.3\%&12&9&9&30.0\%\\
$\pihalf$ Global-FT&B&23&7&0&0&0.0\%&--&--&--&--&--&--&--&--\\
GTI+$\pihalf$&A&26&15&--&--&50.0\% (+6.7)&27&24&23&76.7\% (+63.3)&26&21&21&70.0\% (+40.0)\\
GTI+$\pihalf$&B&27&24&24&12&40.0\% (+40.0)&--&--&--&--&--&--&--&--\\\bottomrule
\end{tabular}
\end{table*}

As shown in Fig.~\ref{fig:rollouts}, we evaluate the AgileX PiPER 6R platform on cabinet search/place, bell-pepper--banana swapping, and kiwi search/place, with representative rollout keyframes. Each condition is independently tested 30 times from the same initial robot pose, with target objects randomized within containers. As shown in Table~\ref{tab:real}, GTI raises aggregate full-task success from 21.7\% to 59.2\% across the three real-robot tasks. Task 1-A improves from 13/30 to 15/30, the four-stage Task 1-B from 0/30 to 12/30, swapping from 4/30 to 23/30, and kiwi from 9/30 to 21/30. Gains concentrate on multi-step conditions; small cabinet handles remain the main cabinet-search failure mode.

\subsection{Runtime of Different System Components}
Runtime is measured for RoboCasa GR00T N1.5 on an NVIDIA RTX PRO 6000 Blackwell Workstation Edition (PyTorch 2.7.0+cu128, CUDA 12.8, BF16, batch 1), after 50 warm-up and 200 timed runs. GR00T predicts 16 actions/call; Qwen uses three views; Wan generates five $256\times256$ frames with 30 denoising steps. Planning runs every three chunks and Wan refreshes only at initialization or confirmed semantic/phase changes. With cached text prompts, Wan averages 1.25 s and a full Planner--Wan--VLA refresh about 1.36 s; low-level execution pauses during refresh. Because Wan is not refreshed every chunk, episode-level effective control frequency depends on the actual number of refreshes. Global-FT has no high-level pause; module-level and end-to-end timings are warmed up independently.

\section{DISCUSSION AND LIMITATIONS}
GTI combines history-based subtask/phase localization, phase-conditioned visual milestones, and failure-triggered local recovery. Improvements on RoboCasa365, VLA-Arena Horizon-Long, and PiPER concentrate on multi-stage tasks. Success still depends on planning, milestone feasibility, low-level interaction, and error correction; GTI complements rather than replaces the VLA.

\textbf{Limitations.} Complex held-out tasks and cross-scene generalization remain difficult. \Kone{}/\Ktwo{} may be too coarse for repeated contact, re-grasping, or mid-course strategy switches. Local recovery also retains the current subtask instead of globally replanning or selecting an alternative strategy. Finer phases, task-level replanning, and stronger interaction policies are natural extensions. 


\end{document}